# Linking Alternative Fuel Vehicles Adoption with Socioeconomic Status and Air Quality Index

### Anuradha Singh[1, 3], Jyoti Yadav[2], Sarahana Shrestha[1, 3], Aparna S. Varde[1, 2, 3]

1. Environmental Science and Management PhD Program
2. Department of Computer Science
3. Clean Energy and Sustainability Analytics Center (CESAC)
Montclair State University, Montclair, NJ 07043, USA
(singha2 | yadavj2 | shresthas1 | vardea)@montclair.edu
ORCID ID: 0000-0002-3170-2510

**Abstract**

Understanding adoption of new technology by its consumers helps us deal with the challenges it faces in any new market. Moreover, the impact it creates on society in terms of the environment, health, and justice is significantly based on the extent of adoption. Alternative fuel vehicles (AFVs) is such an area that faces challenges in terms of consumers' diverse social status and resistance to change. In order to achieve a cleaner transportation sector, we need to address these challenges. In this paper, we conduct a study via machine learning techniques to correlate the adoption of AFVs across various regions, their socioeconomic ranking, and their impact on the air quality index (AQI); and furthermore to predict the AQI as per the region's AFV adoption. This is an empirical study with predictive modeling based on a regional panel data analysis where we use real US census data, air quality data, and data on the number of AFVs purchased per region. Research in this area can help to promote appropriate policies for AFV adoption in the future with due justice to different population groups. This work exemplifies a modest utilization of AI techniques to enhance social good. More specifically, it makes a considerable impact on energy, climate, transportation, and environmental sustainability.

## Introduction

Transportation is as the largest source of greenhouse gas emissions in the United States as mentioned by the Environmental Protection Agency in the literature (EPA 2022). Consequently, both federal and state governments are acting to combat climate change impacts in the country (Bipartisan Infrastructure Law 2021). Analogous to many other states, New Jersey is propagating the use of alternative fuel vehicles (AFVs) including electric vehicles (EVs) to achieve the state's greenhouse gasses (GHG) target reduction and meet its clean energy goals (NJ GWRA, 2020). As of December 2021, there are only 64,307 electric vehicles registered in NJ, hence it is going to be an uphill task to achieve these targets (NJDEP). Innovative methods need to be adopted for decarbonization of transportation (Milovanoff 2020). Previous studies have highlighted the importance of effective policies and socioeconomic factors for AFV adoption (Xue et al. 2021). A few studies on Air Quality Index (AQI) and AFV sales have been conducted (Guo et al. 2020). This motivates our research constituting pilot study in this paper. We define our objectives as exploring AI-based solutions with the following goals.

1. Analyze the link between regional AFV growth and socioeconomic ranking
2. Estimate the correlation between AFV adoption and ambient air quality per region
3. Build a model to predict air quality using data on AFV adoption and socioeconomic census data

Our study is aimed at supporting faster adoption of AFVs and assisting the policymakers' decision-making.

Understanding the link between social ranking and AFV adoption can guide us to address different needs of the population as well as policy selection for a specific transportation mode. Further, the use of AFVs lowers GHG emissions locally and has a direct and immediate impact on the ambient air quality. Our research in this paper on predictive modeling for air quality based on the AFV adoption and socioeconomic census data for specific regions is novel and can guide policymakers to better decisions. The inclusion of five air pollutants ($SO_2$, $O_3$, $NO_2$, $PM2.5$, and $CO$) also makes it novel, as previous research has mostly focused on $PM2.5$. Our present study is focused on the counties in NJ, based upon which our further work shall involve adding more states to make our model more robust. It is also pertinent to note that air quality is not just affected by transportation but by various other factors such as industrial markets / processes and energy generation methods used for a given region. Other challenges in this initiative relate to the limited availability of data for analysis, because the use of AFVs has not reached its full potential. All the results are based on the assumption that there are no big changes in the concerned factors to have any additional impact on air quality apart from increased AFV usage. Incorporating all the different sources of pollution, and their data collection is a related challenge. Furthermore, AFV vehicle adoption is hampered by the lack of adequate consumer awareness as well as the high cost of AFVs compared to conventional vehicles fueled by gasoline and diesel. The question that most vehicle users would ask is: *Why should I pay more?* The answer to that question lies in the fundamental theme of *social good*. It is important to convey to the masses that their additional costs will yield significantly higher benefits with respect to cleaner air, better environmental conditions, and consequently a much healthier life (which will also save medical costs).

This is precisely where our pilot study in this area aims to make a modest contribution. We investigate the role of AI techniques in order to draw correlations between AFV usage and AQI, and predict AQI based on AFV and related factors, so as to highlight the AFV impact on the masses. We hope this work contributes the 2 cents to AI for social good.

## Related Work

For equitable and efficient transportation, policymakers need to identify areas for improvement and actions that can be taken to improve the current system, e.g. encouraging EV Access and affordability through an understanding of the social structure (Fleming 2018). A study to understand the relationship between the market share of electric vehicles and the presence of government incentives, and other influential socioeconomic factors for the US, concluded that electricity prices were negatively associated with EV use while urban roads and government incentives were positively correlated with states' electric vehicle market share (Soltani-Sobh et. al. 2017).

EV studies on the equity aspects have focused on disproportional EV adoption and cost of failing to provide equal access, ranging from disparities in local pollution (Holland et. al. 2019; Ju et. al. 2020), to unfair distribution of public subsidies (Borenstein and Davis 2016), and disparate changes in neighborhood desirability (Henderson 2020, Rice et. al. 2020). Demographic variables, such as income, gender, age, education, and household size have previously been analyzed (Sovacool et. al. 2018; Soltani-Sobh et. al. 2017; Gallagher and Muehlegger 2011; Langbroek et. al. 2016). Research on the adoption of EVs in India included a SWOC (strength weakness opportunity and challenges) analysis and concluded the need for more support from the central government both for research and businesses (Singh et. al. 2021). The proportion of cleaner transport options in the overall market also decides the amount of their impact, in one related study in China it was found that new energy vehicles cannot be considered as an efficient measure to mitigate air pollution since their numbers are not yet significant, rather focus on cleaner energy production method is needed (Su et. al. 2021). Another piece of research in Barcelona and Madrid shared similar results where 40% EV introduction was needed to improve air quality, especially $NO_2$ and CO; the sources of power generation for the region and other emission sources played important roles in the impacts on air quality improvement (Soret et. al. 2014). Inferences on these lines were observed in research using the Community Multi-scale Air Quality model and Weather Research and Forecasting model in Taiwan Li et.al. 2016) and another one using Lagrangian dispersion model used for pollution prediction for scenario analysis of EV introduction for a highway in Milan, Italy (Ferrero et. al. 2016). Various techniques in data mining and machine learning are adapted for predicting environmental parameters, e.g. air quality vis-à-vis health (Varde et. al. 2022), energy demand in the residential sector (Prasad et.al. 2021) and decision support for green data centers (Pawlish et al. 2012). Such techniques along with NLP methods are deployed to assess human interactions / public sentiments on matters pertaining to air quality, public policy and related aspects (McNamee et. al. 2022, Du et.al. 2020, Field et al, 2022, Du et. al. 2016, Kommu et al. 2022). Interesting multidisciplinary approaches are proposed in such works, with opportunities to analyze the diverse data available and utilize it in urban planning for social good.

Other related research (Das et al. 2018), (Puri et al. 2018) (Babicheva et al. 2016), (Radakovic et al. 2022), (Eisner et al. 2011), (Varde et al. 2004), (Zhang et al. 2022) is noteworthy. Yet, the goals of our study have not been achieved in earlier work in a comprehensive manner: analysis of AFV adoption, socioeconomic census, and air quality data together. This constitutes its novelty.

## Hypothesis, Data and Methods

In this study, we emphasize a simple hypothesis: r*egions with higher AFV adoption will have better air quality and lower pollution levels*. In line with this hypothesis, we aim to find understandable and explainable correlations between AQI and AFV adoption, as per specific regions, which we can use for predictive modeling of AQI based on future increases in AFV numbers. This helps to gain more insights into the relationships between alternative fuel vehicles and the corresponding air quality, at present and in the future. Via this study, we hope to make positive impacts thriving on the theme of AI for social good, via its relevance to clean air, sustainable environment, and enhanced transportation, all of which imply good health for the masses.

The regions of focus in this work are various counties in NJ. The datasets used in this study are: NJ Alternate Fuel Vehicle historical county-based report with the types of alternative fuel vehicles sold; socioeconomic census data for NJ counties on population count, education, unemployment, poverty rate, and median household income, as well as air quality data from monitoring stations in NJ. These datasets are for the years 2016-2022. The datasets are correlated with the AFV used in each county, based on the type of the AFV. The number of vehicles is used to calculate the AFV ranking for each county. In terms of air quality data, we focus on 5 major pollutants: $SO_2$, $O_3$, $NO_2$, PM2.5, and CO. For the purpose of this study, the AFVs considered are: Battery Electric Vehicles (BEV), Plug-in Hybrid Electric Vehicles (PHEV), Neighborhood Electric Vehicles (NEV), and Hybrid Electric Vehicles (HEV). Their definitions are annexed to this paper in the Appendix.

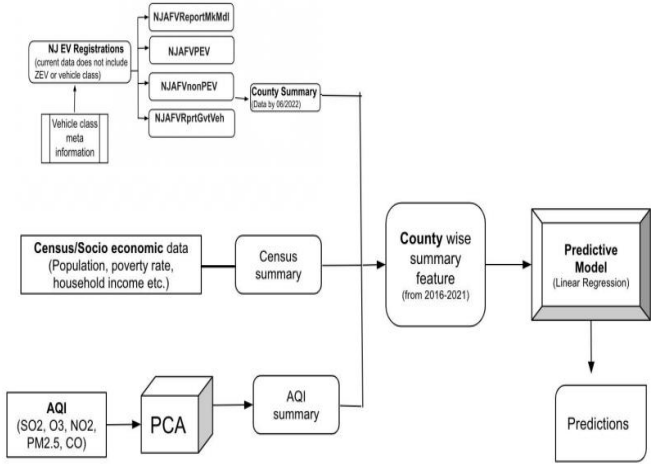

Figure 1: Data Processing Outline

Data preprocessing conducted in this study is illustrated in Figure 1. The NJ Electric Vehicles Registrations dataset is used to obtain county-wise summarized AFV data which includes: semi-annual AFV adoption rate, PEV, non-PEV, and Public Road Mileage and Vehicle Miles Traveled (VMT). In order to prepare the final dataset we compile concatenated AFV data along with socioeconomic data comprising population, poverty rate, and household income. Additionally, we preprocess AQI data by implementing PCA (Principal Components Analysis) on SO2, O3, PM2.5, and CO for dimensionality reduction. Thereafter, we build a predictive model using linear regression to predict AQI using covariates (AFV and socioeconomic data).

We harness Pearson Correlation Coefficient to relate the major variables in socioeconomic data (population, poverty rate, household income, education level) with the extent of AFV adoption. Dimensionality reduction is performed on raw AQI data. Further, the relationship between the AQI scores and these variables is defined by Equation 1.

$$Y_i = \sum_{i=1}^{n} \beta_i X_i + \varepsilon \qquad (1)$$

Here $i$ is the index, $Y_i$ denotes the AQI of the county $i$, each $X_i$ denotes a value while each $\beta_i$ depicts its slope coefficient respectively such that $\varepsilon$ is an approximate error term associated with the equation. We apply this for poverty level, median household income, population of the county and vehicle count in this pilot study. While building our predictive model for AQI, dimensionality reduction is used to avoid overfitting and multi-collinearity. In order to enhance extraction and visualization of relevant data, we perform PCA on the 5 aforementioned pollutants to calculate the AQI for each county. The target variable for prediction is the AQI score.

Linear regression is conducted to predict the AQI via the AFV count and the socioeconomic (SE) factors including median household income, poverty, income, and population. Regression is thereby deployed to capture the relationships between the dependent variable AQI score and independent variables from AFV and socioeconomic data, in order to predict the future values of the target, i.e. AQI. The reason for choosing linear regression in our study is that it provides more understandable and explainable mappings between the input and the output (as opposed to typical deep learning models based on neural networks) that offer black-box approaches, often lacking explainability. In this work it is crucial to fathom how certain factors lead to a given output (and hence make decisions), e.g. how median household income can affect AFV usage and thus AQI.

| Algorithm 1: Predictive model for AQI via SE and AFV |
|---|
| **Input**: D=[σ, α] where σ: SE, AFV variables, α:actual AQI |
| **Parameter**: weights ω, learning rate λ, maximum number of iterations η, error threshold τ |
| **Output**: predicted AQI scores α` |
| 1: Let i = 1, Δ = 100 |
| 2: **while** (i <= η) **or** (Δ < τ) **do** |
| 3:    α` = inner product of σ and ω |
| 4:    Δ = α` − α |
| 5:    γ = 2 dot (σ `, Δ) |
| 6:    ω − = λ * γ |
| 7: **end while** |
| 8: **return** α` |

Algorithm 1 provides the pseudocode for building our AQI predictive model based on learning from existing SE, AFV and AQI data. The experiments are discussed next.

## Experimental Results and Discussion

Experiments are conducted using Python's Scikit Learn. In the results shown here, 80% of the data is used to train the predictive model and 20% of the data is used for testing. Figure 2 plots the semiannual AFV growth by vehicle type.

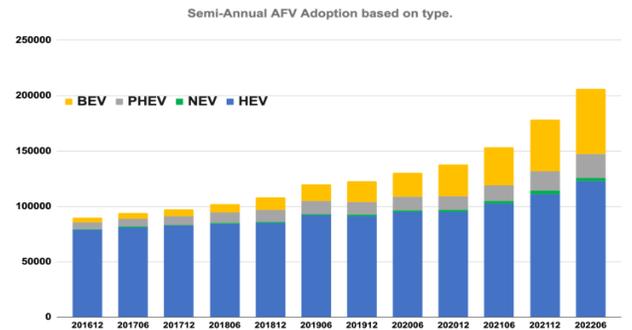

Figure 2: Semiannual AFV growth by type (NJ 2016-2022)

We observe an increase in all types of AFVs across all counties from 2016 to 2020, with a higher increase in the number of Battery Electric Vehicles (BEV) of 15% from 2016 to 2022. The number of Plug-in Hybrid Electric

Vehicles (PHEV) increases by 5.8%, Neighborhood Electric Vehicles (NEV) by 3.8%, and Hybrid Electric Vehicles (HEV) by 1.8%. In total, there are more HEVs, followed by BEVs, PHEVs, and NEVs. The increase in BEVs seems more explainable with the advancements in technology as well as Federal and State subsidies and grants to consumers to uplift their purchase.

Table 1: AFV rank & socioeconomic rank in NJ Counties

| SE Rank | County Name | Population | Total AFV | AFV Rank |
|---|---|---|---|---|
| 1 | Bergen | 953819 | 28304 | 1 |
| 2 | Middlesex | 860807 | 25674 | 2 |
| 6 | Monmouth | 645354 | 19800 | 3 |
| 3 | Essex | 854917 | 19138 | 4 |
| 10 | Morris | 510981 | 16782 | 5 |
| 12 | Mercer | 385898 | 16063 | 6 |
| 8 | Camden | 523771 | 13398 | 7 |
| 11 | Burlington | 464269 | 12862 | 8 |
| 13 | Somerset | 345647 | 12722 | 9 |
| 5 | Ocean | 648998 | 12530 | 10 |
| 7 | Union | 572114 | 11995 | 11 |
| 4 | Hudson | 702463 | 11193 | 12 |
| 9 | Passaic | 518117 | 8755 | 13 |
| 14 | Gloucester | 304477 | 6331 | 14 |
| 15 | Atlantic | 274966 | 6236 | 15 |
| 18 | Hunterdon | 129924 | 5362 | 16 |
| 17 | Sussex | 145543 | 3625 | 17 |
| 20 | Cape May | 95661 | 2888 | 18 |
| 19 | Warren | 110731 | 2814 | 19 |
| 16 | Cumberland | 153627 | 2302 | 20 |
| 21 | Salem | 65046 | 1148 | 21 |

We observe that there is a direct correlation between SE ranking and AFV ranking. Table 1 presents a synopsis of these rankings. However, there are some deviations as well. Bergen County has the highest number of AFVs and is ranked the highest in terms of socioeconomic ranking as well, followed by Middlesex County and Essex County, which showed similar patterns. Salem is the county with the lowest number of AFVs and it also has the lowest socioeconomic ranking. As the socioeconomic status increases, the number of AFVs in the county also increases. Outliers observed here are NJ counties such as Hudson, Monmouth, and Cumberland, where this trend is not found. According to Pearson Correlation Coefficient, the vehicle count, population count, education level, and median household income are found to be significantly correlated with the AFV adoption rate. This further shows that socioeconomic ranking, which accounts for higher median household income, education levels, and population count of the county is found to be directly correlated with AFVs ranking. The highlighted rows show counties with almost the same socio-economic and AFV ranking.

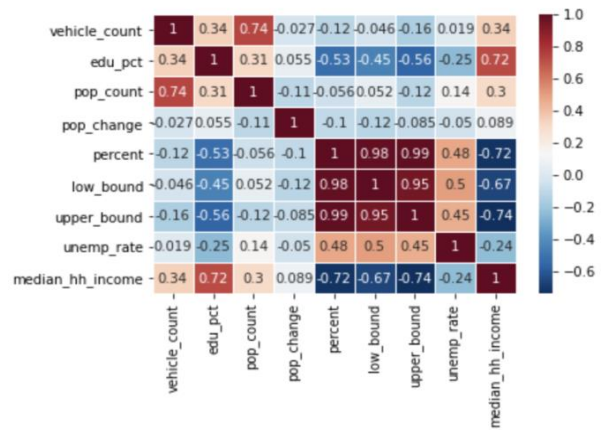

Figure 3: Pearson Correlation Coefficients of socioeconomic census data with AFV ranking

Results of analysis based on Pearson Correlation show observations where socioeconomic factors are positively correlated to AQI. Figure 3 illustrates Pearson Correlation Coefficients linking socioeconomic census data with AFV ranking. Here the variables included are vehicle count, education percentage, population count, population change, percent of population in poverty, the lower and upper bound (gives 90% confidence interval of the poverty percentage), unemployment rate and median household income. The education percentage is highly linked to median household income, and vehicle count is linked with population count.

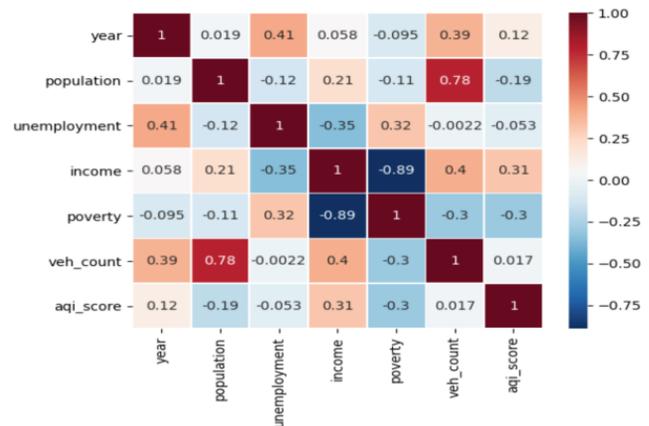

Figure 4: Pearson Correlation Coefficients of AQI and various socioeconomic factors

Figure 4 portrays Pearson Correlation Coefficients linking AQI with socioeconomic factors. Variables observed are the year, population, unemployment, income, poverty rate, vehicle count, and AQI scores for the counties. We observe that AQI scores are linked positively with the income levels and negatively with the population, unemployment, and poverty in the counties.

In our predictive modeling using linear regression, the $R^2$ score is 0.69 and the Mean Squared Error (MSE) is 0.003 over test data. This implies that AQI predictions can occur with high accuracy in order to estimate air quality based on the AFV adoption and socioeconomic factors.

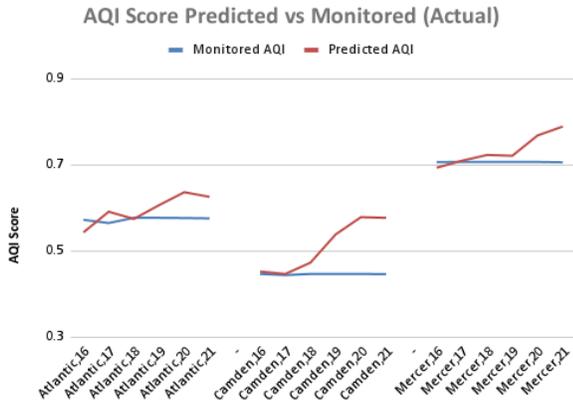

Figure 5: Predictive model for AQI scores indicating a close match between predicted and actual values

Figure 5 plots the results of data tested for 3 NJ counties i.e. Atlantic, Camden, and Mercer over 6 years. These values are listed in Table 2 in the Appendix for a more detailed display. As we can see, the predicted AQI values are in line with the actual AQI for these counties for the years 2016, 2017, 2018, and 2019; however, for the years 2020 and 2021 the predicted values are somewhat higher than the actual monitored AQI values which could be due to the impacts of pandemic and decreased transportation activity overall. This brings our attention to incorporating more variables in future work that shall reflect such changes in transportation systems and their usage. On the whole, observing these 3 years' values, our predictive model seems convincing and can be further enhanced to study the impact that the AFV adoption can have on the air quality of any region.

Based on this entire study, we find that: *there is a lower pollution and better air quality in the counties with a higher number of AFVs and with better socioeconomic ranking*. This confirms our initial hypothesis that AQI scores and AFV counts are positively correlated, and furthermore helps establish the correlation between the socioeconomic ranking and AFVs. Our predictions of AQI are accurate based on the ground truth obtained from existing data, and hence can be used for further estimations on air quality as per AFV usage. Policy decisions for regions that have bad air quality can be undertaken only when the policymakers understand the impact such decisions can have for the given areas. With varied options of AFV and varied socioeconomic strata, our work can take a step towards bringing social equity by helping to choose the best AFVs and public transit options as needed based on regional needs, the respective AQI and related factors such as socioeconomic indicators. Since planning for clean transportation involves consideration of future years, our model to predict AQI score for areas with an estimate of AFV adoption can help governments reach their GHG targets along with social good locally, thus contributing to it initially on a small scale, and in helping to enhance AFV awareness on a larger scale, in the future.

## Conclusions and Roadmap

In the race to mitigate negative effects of climate change, governments and people need to move towards clean energy and clean transportation methods. Clean technologies such as EVs and other AFVs are backbones for future sustainable living, their adoption however needs government support both for businesses and consumers owing to their higher prices as well as limited infrastructure availability. Hence, issues of social equity and social good come to the forefront and government support should not be limited to higher-income consumers alone, considering that the low-income consumers often live in areas with poor air quality. As found in our research study, counties with higher AFVs have better air quality and vice versa. Moreover, those are the counties with high socioeconomic ranking. Our research focusing on predicting the AQI for counties based on their socioeconomic status and AFV adoption is our initial effort to help policymakers' decisions based on social good and to achieve sustainable development goals.

This is a pilot study focusing on the transportation sector and assuming no big changes in the other sectors affect ambient air quality in NJ. Lack of uniform air quality monitored data adds to some of the challenges in this study. Future research in this realm involves adding more data from other states and covering the USA. This study focuses on the use of AFVs which has not yet reached its full potential in NJ and the US. More intensive studies can be conducted on air quality as AFV adoption rises, and we have more data for analyses.

Based on our pilot study, here are a few important takeaways with the scope for future work.
- Policymakers can offer programs in areas with high socioeconomic ranking to encourage more AFV adoption, as these communities can "afford to pay more" for the social good of reducing AQI to improve the environment and be healthier.
- Areas with lower socioeconomic status need more attention because social equity is an important aspect of social good; incentives such as tax cuts, pricing of AFVs proportional to income etc. are needed to proliferate the use of AFVs in order to make them more accessible to low income areas.
- More datasets on AFVs and AQI must be stored with open access to facilitate further AI-based studies; software should be built for the actual AQI prediction based on AFV adoption, to provide at-a-glance information, encouraging more AFV usage.
- Mobile applications (apps) can be developed that link AFVs and AQI, and predict daily AQI levels

to make the masses more aware of air quality and environmental sustainability.
- Further studies with explainable AI models and black-box models should be performed on AFV, SE and AQI data to explore relative merits of the models and use the results for recommendations.
- Models can be introduced such that one of them takes only AFVs adoption into account, while another uses socioeconomic data as well, so as to compare results from a "social good" angle.

In sum, we highlight the fact that AI plays various roles in promoting social good. In our modest study here, machine learning based analyses deploying linear regression and Pearson Correlation Coefficient shed more light on the linkage between AFV adoption, AQI and socioeconomic factors; opening the door to further research, and motivating the enactment of policies to promote more AFV usage. Our paper thus contributes the 2 cents to AI for social good.

## Acknowledgments

Dr. Aparna Varde acknowledges NSF MRI grants 2018575 and 2117308 in the general areas of Big Data and Robotics respectively. She is an Associate Director of the Clean Energy and Sustainability Center (CESAC) at Montclair State University. She is a visiting researcher at Max Planck Institute for Informatics (MPII), Saarbrücken, Germany, ongoing from her sabbatical. Anuradha Singh and Sarahana Shrestha have Doctoral Assistantships from the Graduate School for the College of Science and Mathematics (CSAM) at MSU. Jyoti Yadav has a Graduate Assistantship from the College of Education and Human Services (CEHS) at MSU. We thank our multidisciplinary sources of funding. We thank Dr. Pankaj Lal, Director of CESAC at Montclair State University for his inputs.

# Appendix

Table 2: Predicted and actual AQI scores for NJ counties as obtained from experimental results

| County | AQI Score | Predicted AQI |
|---|---|---|
| Atlantic,16 | 0.573 | 0.543 |
| Atlantic,17 | 0.565 | 0.592 |
| Atlantic,18 | 0.577 | 0.575 |
| Atlantic,19 | 0.577 | 0.607 |
| Atlantic,20 | 0.577 | 0.637 |
| Atlantic,21 | 0.576 | 0.626 |
| Camden,16 | 0.447 | 0.453 |
| Camden,17 | 0.444 | 0.447 |
| Camden,18 | 0.447 | 0.473 |
| Camden,19 | 0.447 | 0.539 |
| Camden,20 | 0.447 | 0.579 |
| Camden,21 | 0.447 | 0.577 |
| Mercer,16 | 0.707 | 0.694 |
| Mercer,17 | 0.707 | 0.710 |
| Mercer,18 | 0.707 | 0.723 |
| Mercer,19 | 0.707 | 0.721 |
| Mercer,20 | 0.707 | 0.768 |
| Mercer,21 | 0.706 | 0.789 |

*Acronyms and Definitions in AFV Terminology*

- **AFV**(Alternative Fuel Vehicle): Vehicle powered by fuels other than Gasoline and/or Diesel exclusively
- **HEV**(Hybrid Electric Vehicle): typically non-plug-in Hybrid Electric Vehicles. Examples: Toyota Prius and many others
- **PHEV**(Plug-in Hybrid Electric Vehicle): typically CARB Transitional Zero Emission Vehicle. Examples: Chevy Volt, Ford C-Max Energi, BMW i3 with range extender
- **BEV**(Battery Electric Vehicle): Examples: Tesla (all models), BMW i3, Nissan Leaf, Chevy Bolt, Honda Clarity
- **PEV**(Plug-in Electric Vehicles): includes both Battery Electric (BEV) and Plug-in Hybrid Vehicles (PHEV), as detailed above.
- **NEV** (Neighborhood Electric Vehicle): otherwise known as Low Speed Vehicles; essentially street-legal golf carts limited to 25 MPH; a type of battery electric vehicle. Examples: ParCar, Columbia, Vantage, GEM
- **NG**(Natural Gas): typically CNG, though may include LNG and propane vehicles (we are unable to differentiate from available data). Examples: Honda Civic, Ford Econoline